\newcommand{\comments}{0}
\newcommand{\anonymous}{0}
\documentclass[conference]{IEEEtran}

\usepackage{cite}
\usepackage{amsmath,amssymb,amsfonts,amsthm}
\usepackage[noend]{algorithmic}
\usepackage{graphicx}
\usepackage{textcomp}
\usepackage{xcolor}
\usepackage{hyperref}
\usepackage{cleveref}
\usepackage{booktabs}
\usepackage{bm} % allowed to add extra packages?
\usepackage{algorithm2e}

\theoremstyle{definition}
\newtheorem{definition}{Definition}
\newtheorem{theorem}{Theorem}

\newcommand{\cM}{\mathcal{M}}
\newcommand{\cR}{\mathcal{R}}
\newcommand{\R}{\mathbb{R}}
\newcommand{\id}{\mathbb{I}}
\newcommand{\gmmsw}{GMMSW~'24}
\newcommand{\asssu}{ASSSU~'26}

\ifnum\comments=1
    \newcommand{\mynote}[2]{{\marginpar{\color{#1}\sf \tiny #2}}}
    \newcommand{\myinlinenote}[2]{{\color{#1} #2}}
    
    \newcommand{\mystrikeout}[2]{{\color{#1} \sout{#2}}}
\else
    \newcommand{\mynote}[2]{}
    \newcommand{\myinlinenote}[2]{}
    
    \newcommand{\mystrikeout}[2]{}
\fi

\newif\iftopicsent
\topicsentfalse % Change to \topicsentfalse to disable
\newcommand{\topicsent}[1]{}

\begin{document}

\title{Fully Node-Private Gradient Descent}

\ifnum\anonymous=1
    \author{}
\else
    \author{
    \IEEEauthorblockN{Colin Bobocea}
    \IEEEauthorblockA{
    \textit{Department of Computer Science} \\
    \textit{Boston University} \\
    Boston, USA \\
    cbobo@bu.edu}
    \and
    \IEEEauthorblockN{Adam Smith}
    \IEEEauthorblockA{
    \textit{Department of Computer Science} \\
    \textit{Boston University} \\
    Boston, USA \\
    ads22@bu.edu}
    }
\fi

\maketitle
\pagestyle{plain}
\thispagestyle{plain}
\begin{abstract}

    Privacy protections in machine learning systems (ML) are typically analyzed assuming a well-defined map from training examples to individuals. However, for many types of data, each example corresponds to several individuals---e.g., an email thread is ``about'' all of its participants---and each individual may contribute many examples.  In such settings, ML models trained with standard, example-level differential privacy (DP) can leak substantial information about individuals. \emph{Node-level} DP, in contrast, asks that a trained model look similar if we remove an individual and all their attributed examples from the dataset. The terminology comes from viewing the dataset as a graph in which nodes are individuals and examples are viewed as edges connecting individuals. Node DP is known to be challenging to achieve for most graph statistics, and there have thus far been no practical ML training algorithms that satisfy full node DP, though several relaxations have been studied. 

    We design and evaluate the first fully node-private algorithm for gradient descent on examples that correspond to edges in a graph. Our algorithm uses, as a key tool, a variant of a recent node-private algorithm for edge-weight estimation (Aden-Ali et al., 2026). We show how this tool can be adapted for gradient estimation in high dimension via \emph{Kashin representations}, and integrated into an end-to-end DP optimization pipeline. We run multiple experiments on two datasets using a logistic regression model and a multi-layer neural network. Our experiments isolate the effects of the parameters of our algorithm, and elucidate the costs incurred for greater privacy. Our work provides an important proof of concept and opens the way to scalable, fully node-private model training.

\end{abstract}


\section{Introduction}
\label{sec:intro}

Differential privacy (DP) \cite{DworkMNS06} has emerged as the de facto standard for protecting the privacy of individuals in the training data of machine learning models, and many works have been dedicated to creating DP optimization algorithms. Models trained with DP algorithms 
provide a crisp, rigorous guarantee against a broad class of attacks, since the presence or absence of a single individual's data does not strongly influence the (distribution of) the trained model's parameters.

In this paper, we study settings where each training example relates to multiple individuals, and each individual can contribute many examples to the dataset. For example, social networks (friendships), messaging and email datasets (sender–receiver pairs), spatial and mobility data (locations visited), transaction records (payer–payee), and co-authorship graphs (shared publications) all have this structure. In such settings, we seek to protect not just any given example, but any given individual and all of their examples. If each example relates to exactly two individuals, we can model the input as a graph in which nodes are individuals, and training examples correspond to edges. The natural notion of privacy is then \emph{node-differential privacy}~\cite{BlockiBDS13,KasiviswanathanNRS13,ChenZ13}. Although a rich theory for node-level DP has been developed, existing work has mostly focused on summarizing the network structure, for example via degree distributions, subgraph counts, or fitting statistical models such as graphons.  

In this work, we consider minimizing a loss function over examples that are attached to edges---for example, fitting a spam classifier to an email dataset. Ganesh, McKenna, McMahan, Smith, and Wu~\cite{IMD2} consider exactly this problem under a relaxation of node-level DP, called \emph{fixed-graph DP} (see \Cref{sec:defs}), which does not provide a guarantee on the privacy of the network structure itself. We show here that, for modest-scale problems, such a relaxation is not necessary: one can get training algorithms with almost exactly the same accuracy and with full node-level protections. 

Concretely, suppose that our dataset is a collection $E$ of triples $(s,t,x)$, where $s,t$ are identifiers of individuals in a set $V$, and $x$ is the training data itself, such as the text of an email thread between $s$ and $t$. In supervised learning settings, $x$ includes a label. We assume that all the training examples for a given pair of individuals have been collapsed into a single record (say, by concatenation); such a grouping does not affect the privacy guarantee. Given a differentiable loss function $\ell$, our goal is to find a parameter $w$ that approximately minimizes $L(w) = \sum_{(s,t,x)\in E}\ell(w;x)$. Example-level DP (equivalently, edge-DP) protects single records, but allows revealing holistic information about an individual's entire collection of records. Treating nodes' edge sets as disjoint datasets (as in standard ``user-level'' DP) is also not sufficient, since it ignores that inserting or removing one node may affect all other nodes' edge sets. Node-DP, on the other hand, ensures an individual and all their records are protected, providing a meaningful guarantee. However, it is challenging to satisfy in practice, largely because it is hard to enforce a contribution bound for individual nodes (e.g. a bound on the graph's maximum degree) while ensuring privacy. 

\subsection{Contributions \& Results}

Our main contribution is an algorithm for node-differentially private gradient descent on a decomposable loss function defined on records associated with edges. Compared to the approach of Ganesh et al.~\cite{IMD2}, which achieves a \emph{weakening} of node-differential privacy, our node-DP algorithm incurs very similar error. Specifically, at the same privacy budget (the $\mu$ parameter of Gaussian differential privacy, \Cref{def:GDP}), our algorithm achieves practically the same predictive performance as the approach of Ganesh et al. across both tasks. Although our algorithm incurs an extra space and runtime cost, it remains practical for modest-scale learning problems.

\paragraph{Technical contribution: Private Contribution Bounding}

The main technical challenge of running node-private gradient descent is to limit the effect of any one node on the gradient at each step. A natural approach is to limit the \emph{maximum degree} of the graph, that is, the maximum number of edges/examples per node. For instance, if all gradient vectors are clipped to a maximum norm of $C$ and the input graph's maximum degree is at most $D$, then the addition of one node can move the gradient at any given round by at most $C\cdot D$. Ganesh et al.~\cite{IMD2} impose a degree bound \emph{nonprivately}: that is, they select a degree-bounded subset of edges (considering only the graph structure, not the content of the training examples), and then run DP gradient descent on the selected subset of examples. This only achieves a relaxation of node-DP since the degree-bounding step can produce vastly different outputs for neighboring input graphs (see, e.g., \cite{RaskhodnikovaS16,DayLL16} for discussion of the failure modes of naive degree-bounding).

We proceed instead by \emph{privately} enforcing a bound on the maximum degree of the input graph.
Given a candidate degree bound $D$ and a privacy parameter $\mu$, our algorithm ensures \emph{node}-level $\mu$-Gaussian DP (\Cref{def:GDP}) on \emph{all} graphs, and enjoys the following accuracy promise: on graphs with maximum degree $D$, the algorithm will exactly run noisy, clipped-gradient descent as if it were \emph{promised} that the maximum degree was $D$. 

At a technical level, we adapt a tool from the recent work of Aden-Ali et al.~\cite{AdenAliSSSU26} (hereafter \asssu{}) and apply it to the setting of high-dimensional gradient descent. They give a quasilinear time algorithm $f_D$ which stably computes the sum of edge weights $w_e \in [0,1]$. In this context, their algorithm is stable in the sense that, if the input graph has maximum degree at most $D$, their algorithm returns the desired sum, $f_D(G) = \sum_{e\in E} w_e$. However, for \emph{any} node-neighboring pair of input graphs $G$ and $G'$ (regardless of their degrees), the values $f_D(G)$ and $f_D(G')$ differ by at most $D$. (Algorithms with such guarantees are called \emph{Lipschitz extensions}; see \Cref{sec:related} for related work.) Using the same algorithm for graphs with edge weights in $[-1, 1]$ and straightforwardly extending the analysis shows that values for node-neighboring graphs can differ by at most $2D$. Instead, we build on their algorithm by processing the graph into its positive and negative weight components, then running the original projection separately and combining the results. This gives a Lipschitz constant of $D$, effectively saving a factor of 2 just through a slight modification (see \Cref{sec:projection}).

We then show how to implement this tool in the high-dimensional setting of gradient descent. At each step, we need to sum vector-valued gradients with bounded Euclidean ($\ell_2$) norms. \asssu{} suggest using their tool coordinate-wise, based on an $\ell_\infty$ norm bound. One way to enforce this is via \emph{Kashin representations} 
\cite{UPandvectorquantization}, originally introduced for signal compression. These allow one to reduce from summing Euclidean-bounded vectors to summing vectors with bounded $\ell_\infty$ norm at a constant increase in sensitivity.

Kashin representations are a family of  nonlinear injective maps from $\ell_2^d$ to $\ell_\infty^{d'}$ whose inverses are linear and whose image dimension $d'>d$ can be chosen to trade off computational efficiency and sensitivity (in a sense we explain in \Cref{sec:method}). We modify the representation to allow for an additional tunable parameter, the $\ell_\infty$ clipping bound. This latter parameter replaces direct $\ell_2$ clipping of the gradient and gives us a tradeoff between bias and variance in the gradient descent step. Our algorithm is node-private regardless of the choice of these hyperparameters; like step size in gradient descent, they only influence accuracy.

\topicsent{This subsection has the following structure. We first introduce the datasets and experiments (accuracy and runtime, varying mu and D). Then, we discuss accuracy results, and we give a brief interpretation of the results. Next, give runtime results and again a short explanation of what's going on. Lastly, we talk about the different parameter choices we can make that effect results (like the kashin params, and we experiment with those as well. A short paragraph at the end puts these results and tradeoffs into context.z}
\paragraph{Experimental Results} 
\topicsent{Describe the setup: the datasets (size) the models (size), what algorithm we compare against, what we test (accuracy and speed)}
We evaluate our algorithm on two text-classification tasks on network-structured data: predicting votes based on their accompanying comments in the Wikipedia Request for Adminship dataset (Wikivotes) using logistic regression, and classifying Reddit comments into discourse categories using a neural network. We compare our algorithm against non-private gradient descent and a full-batch implementation of the fixed-graph DP algorithm of Ganesh et al.~\cite{IMD2} (which we will call \gmmsw{}). We report these results in \Cref{sec:results}. Our aim in running these experiments is to compare the accuracy and runtime of the algorithms. We also seek to gain deeper insights into the inherent tradeoffs of the algorithm by isolating the effects of hyperparameters; namely, the privacy budget, degree bound, and parameters of the Kashin representation.

\begin{figure}[t]
    \centering
    {\small Wikivotes\par}
    \includegraphics[width=\linewidth]{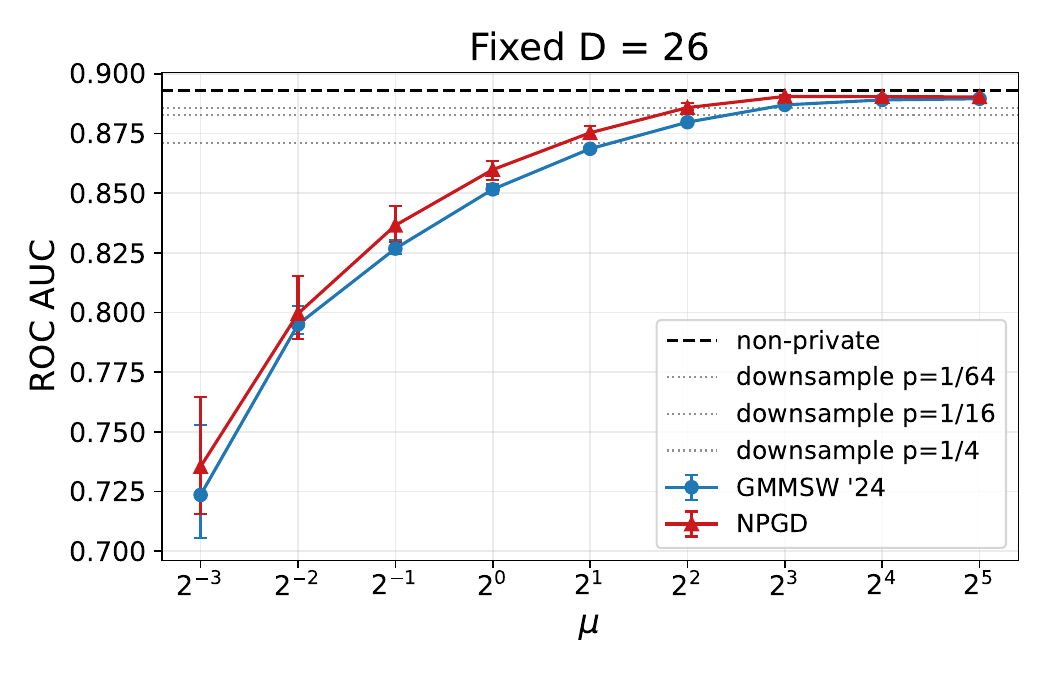}

    \vspace{0.5em}
    {\small Reddit\par}
    \includegraphics[width=\linewidth]{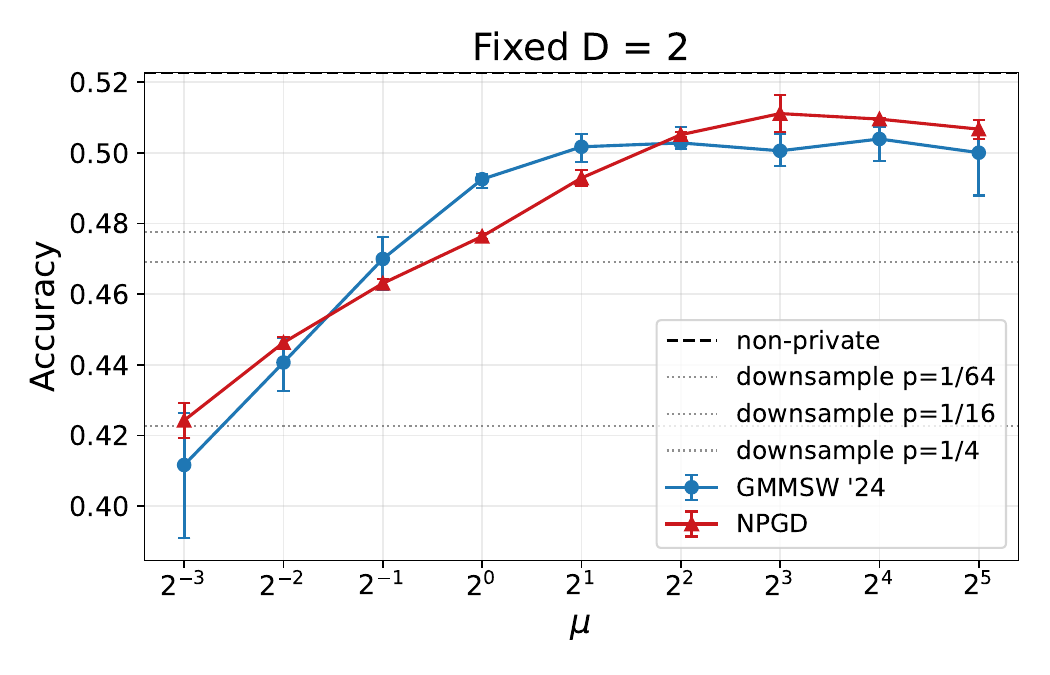}

    \caption{Predictive performance as the privacy parameter $\mu$ varies for models trained on Wikivotes (top) and Reddit (bottom). We report ROC-AUC for Wikivotes and classification accuracy for Reddit, comparing our node-private gradient descent algorithm against the fixed-graph DP algorithm of \cite{IMD2} (labeled ``\gmmsw'') and non-private baselines.}
    \label{fig:vary-mu-performance}
\end{figure}

\topicsent{Discuss results of experiments, point out AUC, say we get very similar accuracy and better privacy}
In our experiments, despite the stronger privacy guarantee of our algorithm, the classifiers produced by our algorithm (NPGD) and \gmmsw{} show very similar predictive performance on both tasks. On the Wikivotes task, we primarily measure performance based on ROC--AUC, as it is known to be a robust measure of performance for binary classifiers. We generally find that NPGD achieves similar (or slightly higher) ROC--AUC than \gmmsw{}, whereas \gmmsw{} achieves lower test loss. This difference is due to NPGD's predictions being less calibrated, tending to be under-confident and thus incurring greater loss, even though their ranking performance is slightly better. We verify this empirically by analyzing logits and calibration curves. On the Reddit task, we report accuracy, though it largely agrees with other multiclass classification metrics such as balanced accuracy and macro F1 score. We again see broadly similar performance across the different metrics we analyze. Both accuracy and test loss are closely matched for the two algorithms across privacy budgets, with neither method showing a substantial advantage (see \Cref{fig:vary-mu-performance,fig:vary-deg}).

\topicsent{The reason the classifiers can differ comes down to a complicated mixture of effects from noise to bias and hyperparameter tuning.}
Though small, we observe some differences in performance metrics for the two algorithms. There is no single, clean explanation for these differences, but rather they arise out of a complex combination of effects including noise, bias, the learning task and loss landscape, and hyperparameter selection. The difference in noise scales is roughly characterized by the Kashin constant $K$, which appears as a multiplicative noise factor and can be reduced through the choice of representation hyperparameters, at the cost of additional computation. The algorithms also introduce different biases: \gmmsw{} fixes a randomly selected subgraph throughout training, whereas NPGD selects different gradient-dependent subsets at each iteration. Lastly, step sizes and clipping norms are tuned separately for each algorithm, further affecting both the magnitude and direction of the resulting parameter updates.
 
\topicsent{price we pay for better privacy is runtime, give numbers and discuss bottleneck}
Because of the additional steps that ensure full node-privacy, our algorithm runs more slowly than the fixed-graph DP algorithm of \gmmsw{}.
%The additional steps in our algorithm increase runtime noticeably, although they remain feasible. 
Averaged over the configurations in the $\mu$ sweep, training a roughly 27,000-parameter model on Reddit, a graph with 50,000 edges, takes approximately $5$ hours with NPGD on an academic cluster, compared with about $9$ minutes for the fixed-graph variant of \cite{IMD2}. 
On Wikivotes, which contains 160,000 edges, training a 384-parameter linear model takes roughly $4$ minutes for NPGD and $11$ seconds for \gmmsw{} (see \Cref{sec:results}). 
There are two main reasons for the increased runtime. First, the Lipschitz extension and Kashin representation require additional computation. Second, because these operations act on rows (Kashin) and columns (Lipschitz extension) in alternation, our algorithm cannot take advantage of batching techniques and incurs more reads from disk. 
%This increase is mostly due to the per-coordinate Lipschitz extensions, though the Kashin representation accounts for a non-trivial portion of the runtime. Running the Kashin representation row-wise (per example), then computing the extension column-wise (per coordinate), increases the number of passes through the dataset and necessitates reading from disk. 

\topicsent{There are also a bunch of parameter choices we can make and ways to tune that affect the alg, detailed here}
Our parameter sweeps also identify tradeoffs that matter for implementing node-private training. For instance, at a fixed privacy budget, increasing the degree bound retains more data but requires more noise. On both tasks, the best predictive performance occurs at degree bounds well below the maximum degree of the input graph. Another tradeoff we observe is the parameter selection for the Kashin representation. Tuning parameters for smaller reconstruction error and smaller dynamic range can improve accuracy, yet comes at the cost of additional computation, with diminishing gains as training time increases (see \Cref{fig:wikivotes-kashin,fig:wikivotes-kashin-kemp}).

\topicsent{wrap up quick word on the implications of these experiments}
Taken together, our results demonstrate that full node-privacy is achievable at largely the same accuracy, but with noticeable costs in runtime. This increase in runtime does not seem fundamental; rather, it appears to be an artifact of our currently limited toolbox of algorithmic techniques. We hope our work spurs investigation into further algorithmic improvements.

\subsection{Definitions}
\label{sec:defs}

The algorithms in this paper satisfy a variant of DP called Gaussian Differential Privacy (GDP)~\cite{DongRS19} that tightly captures the leakage of Gaussian noise addition. As with other notions of DP, it is specified by a neighbor relation, which captures which pairs of datasets are neighbors, and a leakage parameter $\mu\geq 0$.  

\begin{definition}\label{def:GDP}
    Two random variables $X,Y$ defined on the same domain are \emph{$\mu$-close} if they can be realized as the postprocessing of two Normal random variables whose means differ by $\mu$, that is, there exists a measurable, randomized function $F$ such that $F(A)\sim X$ and $F(B)\sim Y$, where $A\sim N(0,1)$ and $B\sim N(\mu,1)$. 

    A randomized mechanism $\cM$ satisfies \emph{$\mu$-Gaussian Differential Privacy} with respect to a relation $\cR$ if, for all pairs of neighboring datasets $G,G'$ (that is, for which $G \sim_\cR G'$), the random variables $\cM(G)$ and $\cM(G')$ are $\mu$-close. 
\end{definition}

Gaussian DP lies ``between'' the commonly considered pure and approximate versions of differential privacy. Namely, every $(\varepsilon,0)$-DP algorithm is $\mu$-GDP for $\mu=O(\varepsilon)$, and every $\mu$-GDP algorithm is $(\varepsilon,\delta)$-DP for $\varepsilon = O(\mu\sqrt{\ln(1/\delta)} + \mu^2)$, for every $\delta>0$; see  \cite{BalleW18,DongRS19} for  more exact conversions.

We use the following two results on GDP. Given a function  $f$ mapping datasets to vectors in $\R^d$ and a relation $\cR$, the $\ell_2$ sensitivity of $f$ is $\text{sens}(f) = \sup_{G,G': G\sim_\cR G'} \|f(G) - f(G')\|_2$. 
\begin{theorem}[\!\!{\cite[Theorem 2.7]{DongRS19}}]
\label{thrm:guassdp}
    The Gaussian Mechanism on a function $f$ taking values in $\R^d$, defined as $\cM(S) = f(S) + Z$ where $Z \sim \mathcal{N}\left(0,  \frac{\text{sens}(f)^2}{\mu^2} \id_d\right)$, is $\mu$-GDP.
\end{theorem}
\begin{theorem}[\!\!{\cite[Corollary 3.3]{DongRS19}}]
\label{thrm:composition}
    The $n$-fold composition of $\mu_i$-GDP mechanisms is $\sqrt{\mu^2_1 + \cdots + \mu^2_n}$-GDP.
\end{theorem}

\subsection{Differential Privacy for Network-Structured Data}
\label{sec:nodeDP}

To use this definition of DP, we have to specify the neighboring relation, which captures the ``unit'' of privacy, that is, what we view as an individual's contribution to the data. 
%Using this definition of DP, the next part to specify is the so-called ``unit" of privacy. That is, what do we consider as one unit of data that changed between neighboring datasets. 
Varying this unit and the %exact 
definition of adjacency determines the differences in privacy notions below. 

Recall that our datasets are graphs $G$ consisting of a collection $E$ of triples $(s,t,x)$, where $s,t$ are identifiers of individuals in a set $V$, and $x$ is the training data itself, such as the text of an email thread between $s$ and $t$. We use $x(e)$ to explicitly denote the third component of the edge $e$. The notions below make sense when each example is attached to an arbitrary set of nodes (creating a hypergraph instead of a graph), not just to pairs; we focus on the graph setting in this paper.

\paragraph{Example/Edge DP \cite{NissimRS07,HayLMJ09}} Here our data unit is a single example in the dataset. We define two neighboring datasets $G$ and $G'$ as differing by the insertion or removal of a single edge along with its associated data. %This guarantees protection for any given example. In our graph setup this corresponds to edge-DP. Meaning, two graphs $G$ and $G'$ are neighboring if they differ on a single edge. 

\paragraph{Node DP \cite{BlockiBDS13,KasiviswanathanNRS13,ChenZ13}} The data unit is a node, and we take two graphs to be neighbors if they differ in the insertion of a node with an arbitrary set of adjacent edges with associated examples. 
This is the notion of privacy we study, since it captures ``user'' (or ``individual'') level DP in the multi-attribution setting, where nodes are people. 
%, in particular we restrict our dataset to have to attributor for each example. We define any two graphs as neighboring if one can be obtained from the other by removing one node and all adjacent edges. 

\paragraph{Fixed-graph DP~\cite{IMD2}} This definition was introduced as a tractable relaxation of node DP. Two graphs are deemed neighbors if they share the same set of nodes and edges and \emph{differ only in the data associated} to the edges of one individual. For example, this allows one to change the content of all of one individual's email threads, but not their set of correspondents. 

%We again consider the user to be our data unit, and adopt the multi-attribution model, however now the exact meaning of neighboring graphs has changed. It says that two graphs are neighboring if for any given node we can arbitrarily change the weights on all its edges. This roughly corresponds to saying that the content of any individual is protected, but the graph structure itself isn't.

%\textbf{User DP.} In user DP, the data unit is a user that contributes examples to the dataset. User DP can be specified in the single-attribution case. Here users can contribute many examples and each example is attributed to exactly one user. Thus, two datasets $S$ and $S'$ are neighboring if one can be obtained from the other by adding or removing a user and all their attributed examples. In the multi-attribution case, we get node DP, described below.

%\cbnote{very unsure as to whether this should be included, its maybe somewhat of a weak example but feels like it would be nice to motivate the need for node DP} 
The discrepancy between fixed-graph DP and node DP is most notable when the graph structure itself contains sensitive information. As an example, consider a messaging platform that trains a model to predict whether two users are likely to communicate. Here, nodes correspond to users, edges represent observed conversations, and edge weights contain features of the messages. An attacker with access to the trained model could query pairs $\{u,v\}$ and rank them by their predicted probability of communication. Under fixed-graph DP, we are guaranteed that changing all message features associated with $u$ must have little effect on the model, but adding or removing $u$ and its incident edges need not. The model may therefore assign noticeably higher scores to pairs that appeared as edges during training, allowing an attacker to infer $u$'s contacts. This attack does not rely on the message contents; it could succeed even if every message were replaced by the same public value.

Node DP provides stronger protection against this attack because it requires models trained on $G$ and on the graph obtained by removing $u$ and all of its incident edges to be statistically indistinguishable. %Since the attacker's queries are post-processing of the trained model, 
The attacker cannot reliably determine whether $u$ and its collection of relationships appeared in the training graph beyond what is allowed by the privacy parameters. This makes reliable link prediction (based on node identities) impossible. Nevertheless, we show that for a rich class of tasks, one can in fact perform accurate machine learning subject to full node-privacy.
%A caveat for this example is that link-prediction task differs from the learning problems we study in this paper, where the model is trained to predict properties of the edge contents rather than the presence of the edges themselves. Nevertheless, it illustrates the distinction between the two definitions: fixed-graph DP does not protect information encoded solely by the graph structure, whereas node DP does.

\subsection{Related Work}
\label{sec:related}

The works below use different variants of differential privacy (pure, approximate, concentrated, R\'enyi, Gaussian, ...). We focus our discussion on the unit of privacy; we do not distinguish the specific notions of distributional closeness.

%\asnote{This should be organized by category a bit differently. I suggest first the IMD2 paper and fixed-graph DP; next the work on user-level DP with multiple records per user (but no overlap); finally, the broader literature on node-DP. I've added paragraph headings. If I get time, I can try to rewrite.}

\paragraph{Fixed-graph Node-level Privacy} The most closely related work is that of Ganesh et al.~\cite{IMD2}, which introduced the problem of training on examples attributed to edges in a graph or hypergraph. Their approach first selects a degree-$D$-bounded subgraph of the input graph---obliviously of the edge data---and then runs an edge-DP algorithm with parameters set to protect any set of $D$ edges. 

Our goal in this work was to see whether fully node-DP algorithms could match the accuracy of their approach.  They implement a follow-the-leader style online optimization algorithm. In our experiments, we instantiate a full-gradient-descent version (which is generally slower but more accurate than the algorithm they implement) of their approach to provide the most fair comparison. 

We do not address the hypergraph version of their open question---see \Cref{sec:discussion} for discussion.
%Our work attempts to answer a question originally posed in \cite{IMD2}. The authors give an efficient algorithm for DP training in the multi-attribution setting, satisfying \emph{fixed-graph} DP. They define this notion of DP to protect all the content of examples a given individual has contributed to the dataset, but not the graph the graph structure itself. In treating the graph structure as public information, they sidestep the difficult issue sidestep the difficult issue of overlapping user contributions, which prevents the guarantee from satisfying full node privacy. This also allows them to consider a more general setting, where examples can be attributed to arbitrarily many users, not just two. \cbnote{Wanted to try to find some stat in IMD2 or somewhere else that said what percentage of their dataset was two-attribution, couldn't find anything but maybe we can just make a general note that a large percentage of these datasets are just two-attribution}. Overall, the paper provides an efficient solution for the general multi-attribution setting, but does so by weakening the privacy guarantee to fixed-graph DP. Our work is motivated by question of whether or not the privacy guarantee does indeed need to be weakened to accurately train models.

\paragraph{Fully Node-Level Privacy for Graph Statistics} Prior work has developed rich techniques for designing node-differentially private algorithms, generally for tasks that directly depend on the graph structure (such as counting subgraphs~\cite{BlockiBDS13,KasiviswanathanNRS13,ChenZ13,RaskhodnikovaS16,DayLL16} and connected components~\cite{KalemajRST25}, and fitting network formation models~\cite{BorgsCSZ18,SealfonU19}). There are generic, exponential-time algorithms that apply to essentially any task with one-dimensional output and any adjacency structure~\cite{FangDY22,LinderRSS25}; these are not practical even at very modest scales. More relevant are a family of general transformations that pre-process the graph to ensure low edge sensitivity~\cite{JainRSS23}; these methods, however, provide poor accuracy when the intended degree bound is very small. (In our larger example, most nodes have degree one or two.) Finally, there is the recent work of \asssu{}, discussed above, which we build on heavily.

\paragraph{User-level Privacy (without overlap among users)} Another line of work explores settings where each individual (``user'') contributes a set of training examples, but those sets are disjoint---each example is attributed to a single individual (e.g., \cite{McMahanRTZ18,AminKMV19,WilsonZLDSG20,LiuSYKR20,LevySAKKMS21,CummingsFMT22,BassilyS23,AsiL24}). As with node-level DP, the main challenge is bounding each individual's contribution to the computation. The required techniques, however, end up being quite different from those that achieve node DP since, in the node DP setting, the insertion of one individual/node can potentially affect all others' datasets.

\section{Methods}
\label{sec:method}

\begin{algorithm}[t]
\caption{Node Private Gradient Descent}
\label{alg:npgd}
\small
\begin{algorithmic}[1]
\REQUIRE Graph $G = (V, E, x)$ with edge-associated examples $x(e)$,
  rounds $T$, learning rate $\eta$, privacy level $\mu$,
  degree bound $D$, Kashin parameters $(\lambda, \delta, \tau, r)$, $\ell_\infty$ clip norm $C_\infty$
\ENSURE Learned parameters $\theta^{(1)}, \dots, \theta^{(T)}$
\STATE $U \gets \textsc{TightFrame}(d, \lambda)$, with output dimension $d'$ 
\STATE Initialize $\theta^{(0)} \in \mathbb{R}^{d}$ with $\theta^{(0)} \sim \mathcal{N}_d(0, 0.01^2)$
\FOR{$i = 1$ \TO $T$}
  \FORALL{$e \in E$}
    \STATE $\bm{g}_{e} \gets \nabla \ell(\theta^{(i-1)}; x(e))$
    \STATE $\hat{x}(e) \gets \mathrm{clip}_{\infty}\!\left(\textsc{Kashin}(U, \bm{g}^{(i)}_{e}, \delta, r, \tau), C_\infty \right)$ \label{line:npdgclip}
  \ENDFOR
  \STATE $\sigma \gets C_\infty D\sqrt{Td'}/\mu$
  \FOR{$j = 1$ \TO $d'$}
    \STATE $G_{j} \gets (V, E, \hat{x}_j)$ \COMMENT{scalar weights from coordinate $j$}
    \STATE $Y_{j} \sim \mathcal{N}(0, \sigma^{2})$
    \STATE $\tilde{\bm{g}}_{j} \gets \textsc{SplitAndSum}(G_j, D)  + Y_{j}$
  \ENDFOR
  \STATE $\bar{\bm{g}} \gets U\tilde{\bm{g}}$
  \STATE $\theta^{(i)} \gets \theta^{(i-1)} - \eta\,\bar{\bm{g}}$
\ENDFOR
\STATE \textbf{return} $\theta^{(1)}, \dots, \theta^{(T)}$
\end{algorithmic}
\end{algorithm}

% \tb{Describe the general flow of our algorithm, we take in inputs similar to normal GD, add privacy level, degree bound, clip norm for privacy and kashin params. Then say the tight frame is required for the kashin rep, once we take the gradients we transform them, then do the projection coordinate-wise, then get the sum of edge weights for each coord and do the update and add noise in the kashin space.}

This section contains a discussion of our algorithm, the key subroutines it relies on, and the guarantees it provides. We modify standard gradient descent by adding Gaussian noise to sums of gradients yielding node-private parameter updates. At each iteration, we start by computing gradients and taking their Kashin representation. The Kashin representation can be thought of as spreading out the information of the original vector into a new, higher-dimensional vector that has small, close to uniform entries. Next, we apply the one-dimensional Lipschitz extension of \asssu{} independently to each coordinate of the transformed gradient vector. This ensures the resulting vector has bounded sensitivity. To ensure node-level $\mu$-GDP of the final parameters, we add noise calibrated to the degree bound $D$, $\ell_\infty$ clip norm $C_\infty$, number of iterations $T$, the transformation dimension $d'$, and privacy parameter $\mu$. Finally, we transform our gradient vectors back into their original space and do the typical GD update. Algorithm~\ref{alg:npgd} summarizes this procedure; the following subsections describe the projection and Kashin representation, compare the resulting noise with fixed-graph DP, and prove the privacy guarantee.

\subsection{Projection} 
\label{sec:projection}
Our method is based on the projection algorithm of \asssu{}. We first present their projection and then show how our variation extends it to a more general setting.  

\newcommand{\cgzo}{\mathcal{G}^{[0,1]}}
\newcommand{\cgoo}{\mathcal{G}^{[-1,1]}}

Let $\cgzo$ denote the set of undirected graphs with edge weights in $[0,1]$, and let $\cgzo_D \subseteq \cgzo$ denote the subset of these graphs with maximum degree at most $D$. Given $G=(V,E,x)\in \cgzo$, where $x:E\to[0,1]$ describes the edge weights, the algorithm selects a degree-bounded subgraph $\widetilde{G}=(V,\widetilde{E})$ on which we compute
\[
f_\text{sum}^D(G) := \sum_{e\in\widetilde{E}}x(e).
\]
If $G \in \cgzo_D$, the algorithm retains every edge, so $f_\text{sum}^D(G)=f_\text{sum}(G):=\sum_{e\in E}x(e)$. On arbitrary input graphs, it guarantees that
\[
| f_\text{sum}^D(G')-f_\text{sum}^D(G) | \leq D
\]
for every pair of node-neighboring graphs $G,G'$. Thus, $f_\text{sum}^D$ is a $D$-\emph{Lipschitz extension} of $f_\text{sum}$ from $\cgzo_D$ to $\cgzo$. We summarize these properties in the following theorem.
\begin{theorem}[\!\!\cite{AdenAliSSSU26}]
\label{thrm:extension}
    There exists a $D$-Lipschitz extension $f_\text{sum}^D$ of $f_\text{sum}$ from $\cgzo_D$ to $\cgzo$ that is computable in time $O\left(|V| + |E|\log |E|\right)$. 
\end{theorem}
\begin{proof}
    See \cite{AdenAliSSSU26} Theorem 4.2. 
\end{proof}

This result is optimal (that is, the Lipschitz constant cannot be improved) for $[0,1]$ weights. However, we require a method that works for graphs $\cgoo$ with edge weights in $[-1,1]$. The most direct modification of the \asssu{} algorithm to work on graphs in $\cgoo$ (which simply translates and rescales the weights) yields a $2D$-Lipschitz extension. Thus, this approach ``loses'' a factor of 2 relative to the best possible Lipschitz extension; when used inside noise-addition algorithms, this increases the standard deviation of the noise and the error of the optimization procedure by a factor of 2. 

We show that this loss is unnecessary. We modify the \asssu{} algorithm to get an extension with (optimal) Lipschitz constant $D$. The main idea is to split edges into those with positive and negative weights and run one instance of the algorithm on each part. By adjusting the order in which each instance considers edges, we get an algorithm with no blow-up in the Lipschitz constant.

The starting point for our approach is a tighter analysis of the \asssu{} algorithm. Specifically, we show that when edges are considered in decreasing order of their $[0,1]$ weights, we have
\[
0 \leq f_\text{sum}^D(G')-f_\text{sum}^D(G) \leq D
\]
for arbitrary graphs in $\cgzo$. The proof is given in \Cref{appendix:projection}. 

To take advantage of this, we split the graph into two subgraphs, one containing only positive-valued edge weights and the other negative. For the subgraph with positive weights, we apply the \asssu{} projection. For the subgraph with negative weights, we apply the same projection to the absolute values of the weights, and then multiply the output by -1. This is equivalent to sorting the negative-edge weights by \textit{increasing}, instead of decreasing, weight. Finally, we add the results of the two instances. This approach ensures that the overall change due to the addition of a node is at most $D$, because only one of the two outputs we sum can move up, and only one of them can move down. 
% Now, we take the absolute value of the graph with negative-valued edge weights and we get two graphs in $\cgzo$. At this point, we apply the projection and the guarantees of the stronger bound above, and taking the sums of both projected graphs' edge weights gives us a function that is $D$-Lipschitz on $\cgoo$. 
Formally, we obtain:
\begin{theorem}
    \label{thrm:betterextension}
    There exists a $D$-Lipschitz extension $f_\text{split}^D$ of $f_\text{sum}$ from $\cgoo_D$ to $\cgoo$ that is computable in time $O\left(|V| + |E|\log |E|\right)$.
\end{theorem}
\begin{proof}
    See \Cref{appendix:projection}.
\end{proof}
The full pseudocode for the algorithm and a self-contained proof are given in \Cref{appendix:projection}. These proofs also straightforwardly extend to the setting where edge weights are in $[-C, C]$ by simply inserting a factor of $C$ in the Lipschitz constant.

%This implies that for node neighboring graphs we have bounded sensitivity of the sum of edge weights. In contrast to user DP in the single-attribution setting, this stronger privacy notion incurs the cost of an extra factor of 2. When using this Lipschitz-extension argument, this factor of 2 is fixed, unlike the Kashin constant $K$, which can be reduced through the choice of representation parameters. Removing the factor of 2 would require significantly new ideas. 

We note that the coordinate-wise projections may retain different numbers of edges. Dividing every coordinate by a common denominator would therefore be inaccurate, while estimating a separate denominator for each coordinate would consume additional privacy budget. We instead omit the normalization and use the sum of the retained gradients directly. The learning rate controls the effective scale of this sum, so a well-specified learning rate produces the same form of update without requiring private estimates of the projection sizes.

\subsection{Kashin Representation}
\label{sec:kashin-rep}

Despite the factor of 2 improvement, there are still a few key issues we must resolve before being able to adapt the Lipschitz extension to our setting.
First, the Lipschitz extension applies to scalar edge weights $x(e)\in[-C,C]$, whereas each gradient $\bm{g}_e$ is a vector in $\mathbb{R}^d$. An $\ell_2$ bound alone does not provide a sufficiently small coordinate-wise bound: the assumption $\|\bm{g}_e\|_2\leq C_2$ only implies that each coordinate is at most $C_2$ in absolute value.
Directly clipping every coordinate at a substantially smaller threshold, however, may severely distort gradients whose mass is concentrated in a few coordinates. We therefore first compute a Kashin representation, first introduced by Kashin \cite{Kashin1977}, which spreads the gradient across a higher-dimensional space and allows us to enforce a small $\ell_\infty$ clipping threshold with less distortion. Lyubarskii and Vershynin provide an efficient construction for Kashin representations \cite{UPandvectorquantization} that we will use. We describe this representation and its guarantees in the following subsection.

Our algorithm's privacy guarantees do not rely on the properties of the Kashin representation; we enforce privacy via $\ell_\infty$ clipping (see Algorithm~\ref{alg:npgd}, Line~\ref{line:npdgclip}). Nevertheless, the algorithm's accuracy depends on the representation's mathematical structure, which we now describe.

Formally, the Kashin representation is a nonlinear map $g\colon\mathbb{R}^d\to\mathbb{R}^{d'}$, where $d'=\lambda d$ for $\lambda>1$, that encodes the information of the original vector in a higher-dimensional space. Decoding this vector is accomplished using a matrix $U\in\mathbb{R}^{d\times d'}$ with orthonormal rows (called a \textit{tight frame}). 
The transformation is specified by $U$, which determines $d'$ and $\lambda$, and two additional parameters, $\delta$ and $\tau$, in $(0,1)$.

For an appropriate choice of the frame $U$ (see the assumptions of  \cite[Theorem~3.5]{UPandvectorquantization}), the algorithm returns coefficients $g(x)\in\mathbb{R}^{d'}$ and a residual $e\in\mathbb{R}^d$ with the following guarantees:
\begin{enumerate}
    \item The coefficients reconstruct $x$ up to the residual:
    \[
        x = U g(x) + e,
    \]
    where $\|e\|_2/\|x\|_2\leq \tau$. 
    \item The coefficients have bounded dynamic range:
    \[
        \|g(x)\|_\infty
        \leq \frac{K_{U,\delta}}{\sqrt{d'}}\,\|x\|_2.
    \]
    % \item Given a prescribed reconstruction tolerance $0<\tau<\|x\|_2$,
    % \[
    %     \|e\|_2\leq\tau
    %     \;\; \text{after}\;\;
    %     r=O\!\left(\log\frac{\|x\|_2}{\tau}\right) \text{ iterations}.
    % \]
\end{enumerate}
The parameter $\delta\in(0,1)$ influences the overall run time of the algorithm and the scaling constant $K_{U,\delta}$. As $\delta$ approaches 1, the run time increases (roughly as $1/(1-\delta)$) and the factor $K_{U,\delta}$ decreases towards some constant $K_{U,1} > 1$ (which depends on $U$). The parameter $\tau$ influences mostly the run time, which grows with $\log(1/\tau)$. The effect of $\tau$ is small, and we assume in the explanation below that $\tau=0$ for simplicity.

At a high level, the algorithm of \cite{UPandvectorquantization} iteratively computes $y = U^{T}x$, then clips it in $\ell_\infty$, and repeats the same process with the residual vector, i.e. the difference between the original and the reconstruction of the clipped vector. The algorithm is given in Appendix \ref{appendix:kashin-rep} for completeness.

We next use the coefficient bound to quantify the noise introduced by our algorithm. Suppose the original gradient vectors satisfy $\|x\|_2\leq C_2$, and write $K=K_{U,\delta}$. The second guarantee above gives
\[
    \|g(x)\|_\infty
    \leq \frac{KC_2}{\sqrt{d'}},
\]
so we may take $C_\infty=KC_2/\sqrt{d'}$. The coordinate-wise \asssu{} projection has sensitivity $C_\infty D$ in each of the $d'$ transformed coordinates. The $\ell_2$ sensitivity of the vector of projected values, which determines the scale of the Gaussian noise, is therefore
\begin{equation}\label{eq:kashin-sens}
    C_\infty D\sqrt{d'}=C_2DK.    
\end{equation}
When we apply $U$ to reconstruct the noisy sum of the gradients, the noise scale is unaffected: because $U$ has orthonormal rows, it maps a random variable drawn from $N(0,\sigma^2\id_{d'})$ to one distributed as $N(0,\sigma^2\id_d)$ (for any $\sigma>0$).

By comparison, the fixed-graph algorithm of \cite{IMD2} produces a gradient sum at each step whose  $\ell_2$ sensitivity is $C_2D$. Thus, under the Kashin guarantee, and assuming clipping parameters are chosen correctly, the noise scale required by our algorithm (\Cref{eq:kashin-sens}) is a factor of $K$ larger. 

By choosing the $\lambda$ sufficiently large and $\delta$ close to 1, we may get the factor of $K$ arbitrarily close to 1, but no smaller. (The case $K=1$ corresponds to the optimal dynamic-range scale $C_2/\sqrt{d'}$, in which the gradient mass is distributed as uniformly as possible across the transformed coordinates.)

The choices of redundancy $\lambda=d'/d$ and parameter $\delta$ create a tradeoff between the attainable value of $K$ and computational cost. Parameter choices that produce a smaller $K$ can require a higher-dimensional representation or more iterations of the conversion algorithm, increasing both memory use and runtime. We examine this tradeoff empirically in \Cref{fig:wikivotes-kashin}.

The calculation describes above the ideal theoretical behavior. The results of \cite{UPandvectorquantization} show that suitable random frames satisfy the uncertainty principle with high probability, but the stated parameter bounds involve unspecified absolute constants and do not directly provide a usable value of $K$ for a realized finite-dimensional frame. We therefore do not rely on the theoretical coefficient bound to enforce privacy. Instead, we explicitly clip the transformed gradients at a chosen threshold $C_\infty$. 

To get a sense of how well the Kashin representation's properties play out in practice, in some experiments, we record the maximum observed $\ell_2$ norms of gradients and maximum $\ell_\infty$ norms of their representations in runs of nonprivate gradient descent.
We then report the corresponding empirical value
\[
K_\text{emp} = \frac{C_\infty \sqrt{d'}}{C_2}
\]
assuming the clipping norms are set to these observed maxima. 
This quantity gives the exact multiplicative noise factor $K_\text{emp}$ relative to the fixed-graph baseline for the selected clipping thresholds. \Cref{fig:wikivotes-kashin-kemp} in Appendix \ref{appendix:kashin-experiments} reports $K_\text{emp}$ alongside ROC-AUC and runtime for a range of Kashin configurations.

\subsection{Privacy Analysis}
We present the following theorem as our privacy guarantee. 
\begin{theorem}
    Algorithm \ref{alg:npgd} satisfies node-level $\mu$-GDP.
\end{theorem}
\begin{proof}
    Assume that the frame $U$ and all hyperparameters are fixed independently of the private data. Conditional on the outputs of all preceding iterations, the current parameters $\theta^{(i-1)}$ are fixed. For each transformed coordinate, \textsc{SplitAndSum} has sensitivity at most $C_\infty D$ by \Cref{thrm:betterextension}, after scaling the clipped edge weights from $[-C_\infty,C_\infty]$ to $[-1,1]$. The vector of projected gradient sums therefore has $\ell_2$ sensitivity at most $C_\infty D\sqrt{d'}$.

    At every iteration, we add isotropic Gaussian noise with standard deviation
    \[
        \sigma=\frac{C_\infty D\sqrt{Td'}}{\mu}.
    \]
    By Theorem~\ref{thrm:guassdp}, each iteration is therefore $\mu/\sqrt{T}$-GDP. Invoking the composition guarantee in Theorem~\ref{thrm:composition} shows that the full sequence of $T$ iterations is $\mu$-GDP. Finally, decoding the noisy transformed gradient with $U$ and performing the gradient-descent update are post-processing operations and incur no additional privacy loss. Thus, Algorithm~\ref{alg:npgd} is $\mu$-GDP.
\end{proof}

\section{Empirical Results}
\label{sec:results}

This section presents our experimental findings. Through these experiments we characterize the privacy-utility tradeoff controlled by $\mu$ and the balance between projection bias and privacy noise controlled by $D$. We then examine how the Kashin representation affects utility and runtime, and conclude with the implementation choices and optimizations used to make training computationally feasible.

\subsection{Datasets and Learning Tasks}
\paragraph{Wikivotes}
The first dataset is Wikipedia Requests for Adminship, which we refer to as Wikivotes \cite{wikivotes}. Each time a Wikipedia user attempts to become an administrator, other users submit votes to approve or oppose the request. They then describe their reasoning with a short comment (a few sentences or less). Our graph model represents users as nodes; a vote cast by one user on another user's request forms an edge whose associated data are the comment text and the vote itself. Wikivotes has about $10{,}900$ users, $160{,}000$ examples, a mean degree of $29$, a median degree of $4$, and a maximum degree of $1{,}260$. The distribution of votes is about $70\%$ approval and $30\%$ disapproval. Our learning task is binary classification of each vote as approve or oppose using the vote comment as features. We encode each comment using the pretrained \texttt{all-MiniLM-L6-v2} Sentence Transformer \cite{reimers2019sentencebert,allminilml6v2}, which produces a 384-dimensional vector embedding. We train a simple logistic regression model with 384 parameters using binary cross-entropy loss.

\paragraph{Reddit}
The second dataset consists of interactions from the social media platform Reddit \cite{zhang2017characterizing}, and we refer to it as Reddit throughout. Users comment on other users' posts or comments, and the dataset provides a label for the type of comment. There are 9 categories: \textit{answer}, \textit{elaboration}, \textit{question}, \textit{appreciation}, \textit{agreement}, \textit{disagreement}, \textit{humor}, \textit{negative reaction}, and \textit{other}. This dataset suggests a graph with users as nodes, edges between users who have interacted, and edge-associated data consisting of the comments and their categories. The dataset contains about 54,000 users and 50,000 examples, with a mean degree of 1.8, a median degree of 1, and a maximum degree of 176. We aim to predict the category of a comment from its text. There is some class imbalance: $40\%$ of comments belong to the ``answer'' category, three other categories each contain about $13\%$ of the comments, and the remaining categories are in the single digits. We use the same Sentence Transformer as before to encode comments numerically. Then, we use a fully connected neural network that maps each 384-dimensional comment embedding through hidden layers of widths 64 and 32 with ReLU activations, followed by a nine-dimensional output layer. The model has 27,017 trainable parameters and is trained using multiclass cross-entropy loss.

\subsection{Experimental Setup}

\paragraph{Baseline Algorithms} 
We compare our algorithm, NPGD, with several non-private variants of GD and with the algorithm of \cite{IMD2} to assess its performance and isolate the effects of projection and privacy noise. Specifically, we compare to the following baseline algorithms: 

\newcommand{\baseline}[1]{\smallskip\emph{#1}.}

    \baseline{Nonprivate} Fully non-private gradient descent with a fixed learning rate.
    
    \baseline{\gmmsw{}} We implement a variant of the algorithm described in \cite{IMD2}, which we call \gmmsw{}. Ganesh et al.~\cite{IMD2} take a two-phase approach. First, they use a contribution bounding to select a subset of examples whose attribution graph has bounded degree. Among the variants they propose, we recreate their greedy contribution bounding algorithm with no duplicates \cite[Algorithm 2]{IMD2}. Second, they apply a DP optimization procedure (DP-FTRL) on the selected subset. In our experiments, we use a full-batch adaptation of their DP-SGD approach; although it is slower, it generally achieves better predictive performance than DP-FTRL, and more closely aligns the optimization procedure with NPGD. Comparing NPGD with \gmmsw{} therefore illustrates the empirical cost of strengthening the guarantee from fixed-graph privacy to node-level privacy.
    
    \baseline{Nonprivate subgraph} A non-private algorithm that runs the same contribution bounding step as \gmmsw{} before the update loop begins. This procedure is nearly identical to \gmmsw{} without privacy noise, allowing us to distinguish the bias introduced by the projection from the effect of the added noise. 
    
    \baseline{Nonprivate downsampled} Non-private gradient descent, trained on a random subset of our dataset. This down-sampled dataset consists of $pN$ examples for $p \in \{1/4, 1/16, 1/64\}$, where $N$ is the size of the original dataset. These baselines provide an intuitive reference point; for example, we can determine whether a configuration performs comparably to a classifier trained on a random quarter of the data.

%\cbnote{where to put this, just kashin rep info really, maybe in plot captions?}
\paragraph{Kashin Representation Parameters}
For Wikivotes, we use Kashin parameters $\delta=0.95$, $\lambda=5$, and $\tau=10^{-8}$. This more computationally demanding configuration remains tractable for the smaller model. For Reddit, we instead use $\delta=0.5$, $\lambda=2$, and $\tau=10^{-4}$, which keeps the maximum coefficient relatively small without incurring the substantial runtime cost of a larger redundancy factor. The motivation for these choices is discussed further in the Kashin accuracy--runtime analysis in \Cref{sec:kashin_compare}.

\paragraph{Hyperparameter Tuning}
Firstly, we preface this section by stating that we select hyperparameters using non-private exploratory runs. This tuning procedure is not differentially private, and the privacy guarantees stated in this paper apply to each training run conditional on its chosen hyperparameters rather than to the full tuning pipeline. Our objective is to better understand these algorithms, compare their relative performance with the baselines, and elucidate the tradeoffs among privacy, accuracy, and computation, rather than to obtain the best possible predictive performance. This is also a common approach in the literature. In a deployment requiring end-to-end privacy, the hyperparameters would instead need to be fixed in advance, selected using public data, or tuned with an additional privacy budget. 

Given a privacy parameter $\mu$ and degree bound $D$, we first tune the clipping threshold $C_\infty$ so that only a small fraction of transformed-gradient entries are clipped. To do so, we run the algorithm without privacy noise, while retaining the Kashin representation, and record the maximum coefficient magnitude using half of the iterations our full training runs use. One can choose to use this maximum or, as we do, take the $95$th percentile of the recorded values to reduce sensitivity to unusually large gradients early in training. We use an extremely similar procedure for tuning the clip norm $C_2$ used in \gmmsw{}, instead tracking the $\ell_2$ norms encountered. For learning rate tuning, we simply evaluate the loss over a grid of learning rates and select the one that minimizes the loss. We also choose candidate values of $D$ non-privately, guided by the observed node-degree distributions, solely to study how the method performs across different degree bounds rather than to propose an optimal selection rule. Related work studies private selection of degree or sensitivity bounds for graph statistics~\cite{RaskhodnikovaS16,KalemajRST25}.

\subsection{Utility Experiments}

\begin{figure}[t]
    \centering
    % --- Column headers ---
    \begin{minipage}{0.48\linewidth}
        \centering\small Wikivotes
    \end{minipage}\hfill
    \begin{minipage}{0.48\linewidth}
        \centering\small Reddit
    \end{minipage}

    \vspace{0.5em}
    % --- Row 1: ROC-AUC and accuracy---
    \includegraphics[width=0.48\linewidth]{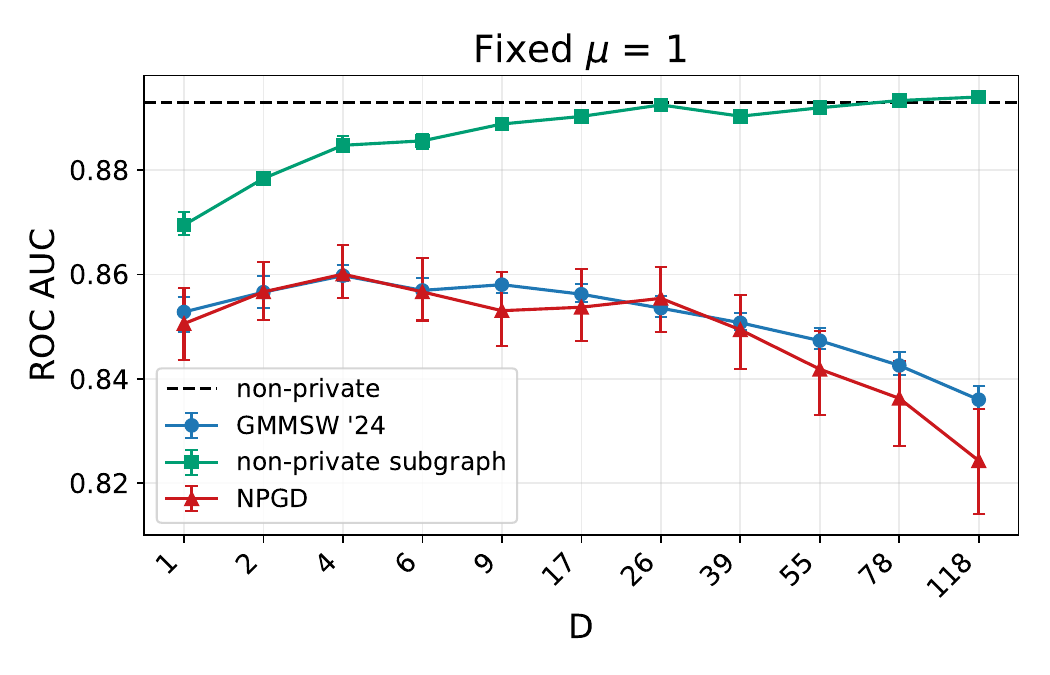}\hfill
    \includegraphics[width=0.48\linewidth]{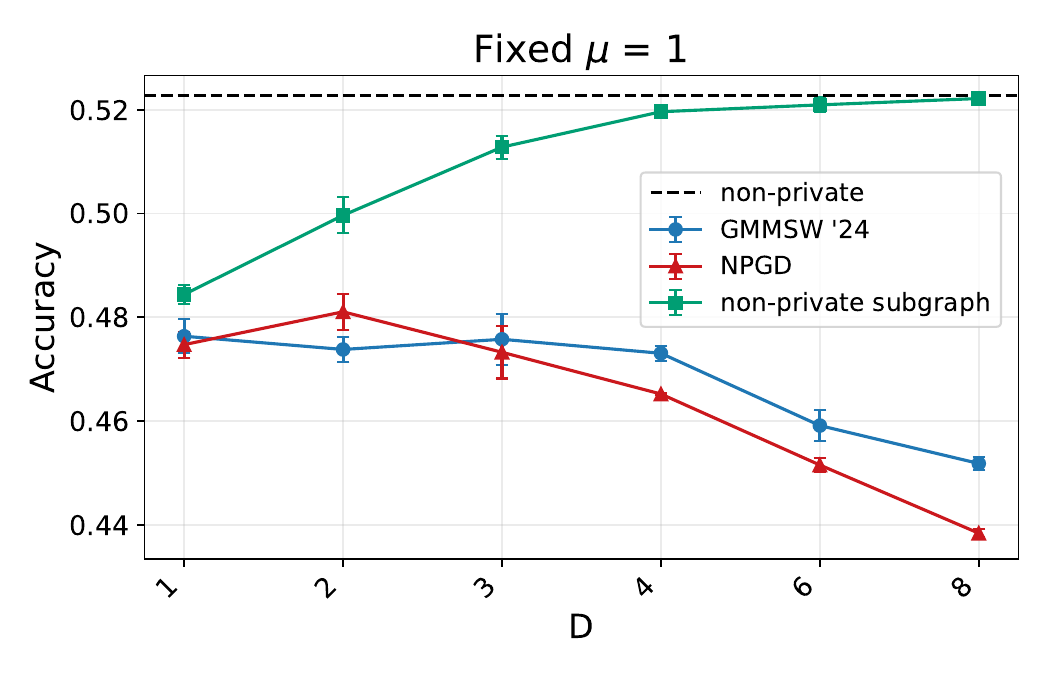}

    % --- Row 2: Test loss ---
    \includegraphics[width=0.48\linewidth]{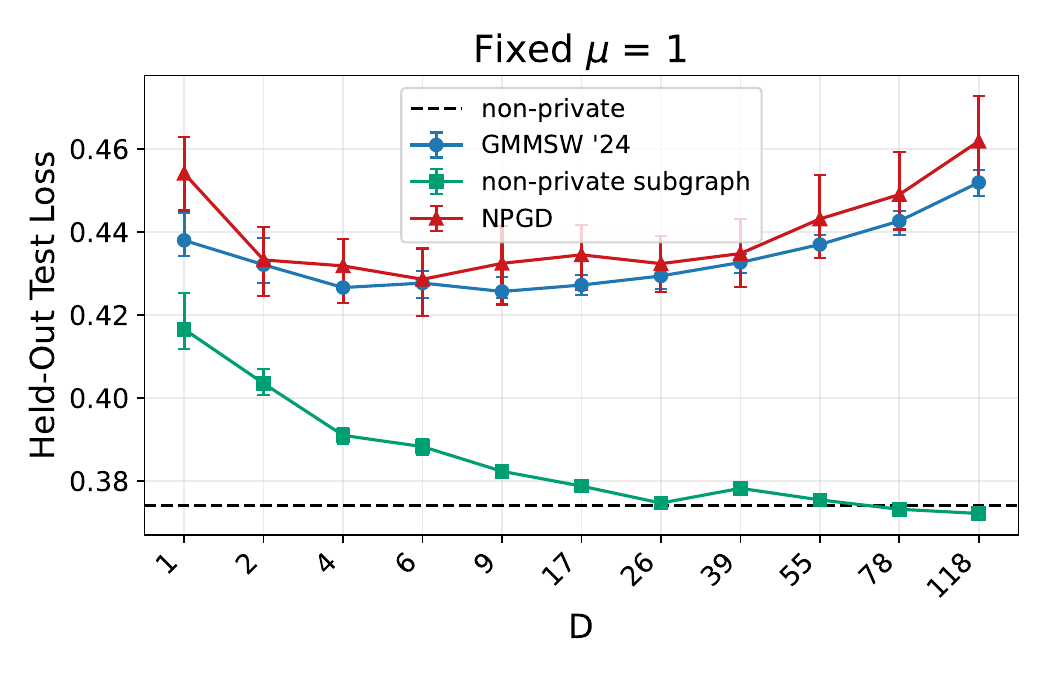}\hfill
    \includegraphics[width=0.48\linewidth]{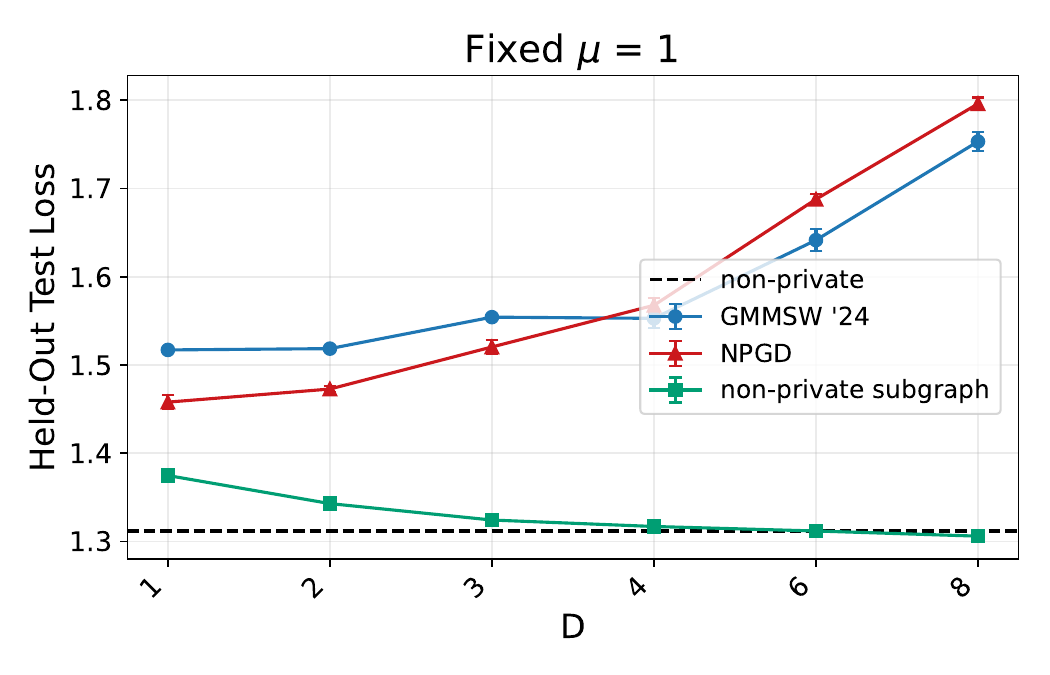}

    \caption{Performance of our algorithm against \gmmsw{} and non-private variations on both Wikivotes (left) and Reddit (right). ROC-AUC (top left) and accuracy (top right) are reported for Wikivotes and Reddit, respectively, and the bottom row shows test loss as we vary the degree bound $D$.}
    \label{fig:vary-deg}
\end{figure}

These experiments examine two central tradeoffs. First, we vary $\mu$ to characterize the privacy--utility tradeoff. Second, we vary $D$ to characterize the balance between projection bias and privacy noise.
\paragraph{$\mu$ Sweep}
Our first experiment fixes the degree bound $D$ while varying the privacy parameter $\mu$ to measure its effect on performance. We select $D=26$ for Wikivotes because it is close to the dataset's mean degree and gives competitive performance across metrics, including the best F1 score in the degree sweep. For Reddit, we select $D=2$ because it is close to the mean degree and gives the best observed NPGD accuracy in the degree sweep. These choices provide representative operating points rather than claiming to be optimal for every metric (see \Cref{fig:vary-deg}). We run the algorithm over the grid $\mu=2^k$ for $k\in\{-3,-2,\dots,5\}$. As $\mu$ increases, the privacy guarantee weakens and less noise is added, so predictive performance should improve. For Wikivotes, we use a Kashin configuration optimized for accuracy ($\delta=0.95$, $\lambda=5$, and $\tau=10^{-8}$). Because training on Reddit is more computationally intensive, we instead use parameters that balance accuracy and efficiency ($\delta=0.5$, $\lambda=2$, and $\tau=10^{-4}$). We are particularly interested in performance in the strong-privacy regime, where $\mu$ is small, and in the comparison with \gmmsw{} at each value of $\mu$.

\begin{figure}[t]
    \centering
    \begin{minipage}{0.48\linewidth}
        \centering\small Wikivotes
    \end{minipage}\hfill
    \begin{minipage}{0.48\linewidth}
        \centering\small Reddit
    \end{minipage}

    \vspace{0.5em}
    \includegraphics[width=0.48\linewidth]{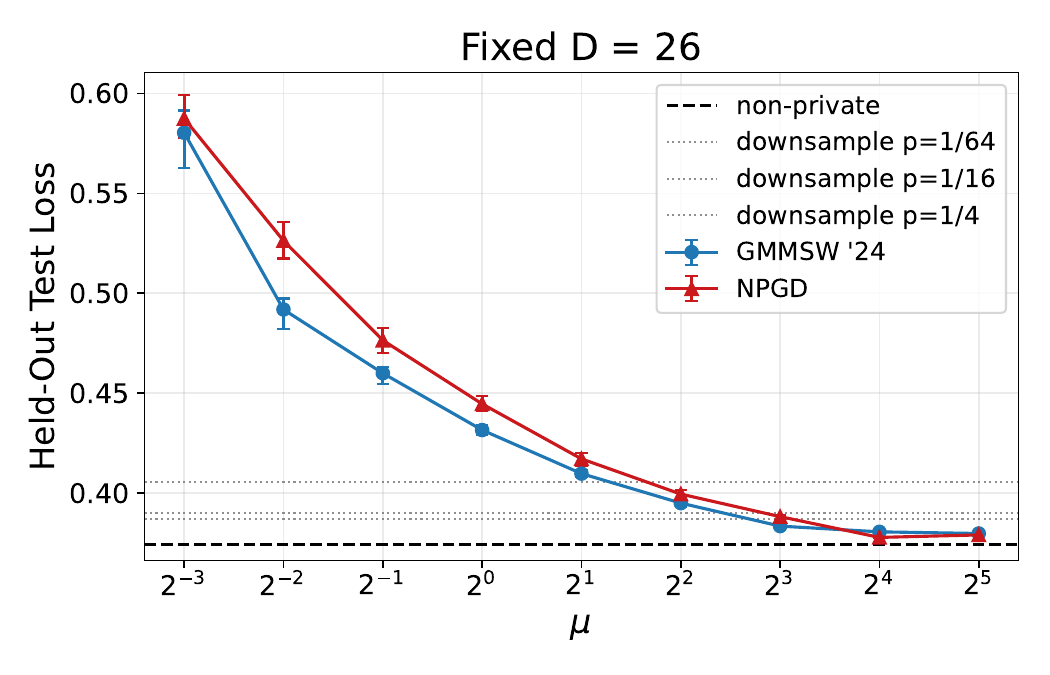}\hfill
    \includegraphics[width=0.48\linewidth]{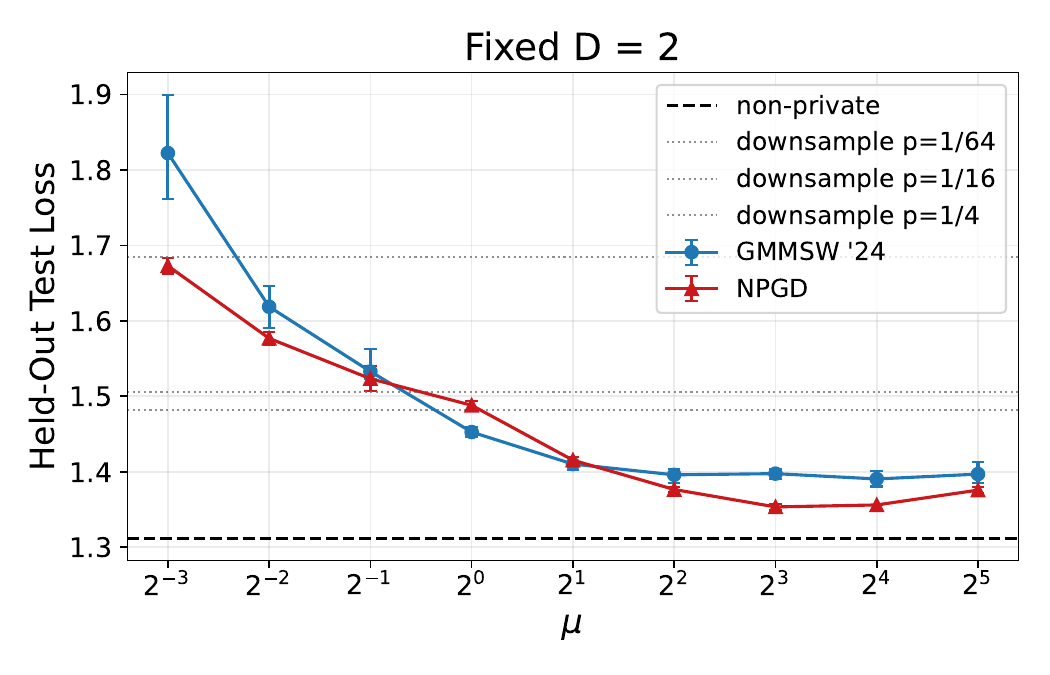}

    \caption{Test loss as the privacy parameter $\mu$ varies for models trained on Wikivotes (left) and Reddit (right).}
    \label{fig:vary-mu-loss}
\end{figure}

The results are summarized in \Cref{fig:vary-mu-performance,fig:vary-mu-loss}. Across both datasets, utility generally improves as $\mu$ increases, consistent with the reduction in noise. On Wikivotes, NPGD has slightly higher ROC-AUC at each $\mu$, while \gmmsw{} generally has lower test loss until $\mu$ becomes large. ROC-AUC measures ranking, while cross-entropy also reflects how well probabilities are calibrated. NPGD can therefore rank examples better without achieving lower loss. The gaps narrow as $\mu$ increases, and both methods approach the non-private baseline. On Reddit, NPGD has slightly better metrics, though it's possible the small gaps are due to run-to-run variability and hyperparameter tuning rather than meaningful differences in performance. On this harder, imbalanced nine-class task, test loss captures improvements in class probabilities that may not change the most likely class, making it more informative than accuracy alone. Both methods improve substantially with $\mu$, while model and optimization error together with projection bias explain the remaining gap to the non-private baseline.

\paragraph{$D$ Sweep}
Next, we reverse the experiment by fixing $\mu=1$, which provides a reasonable balance between privacy and utility, and varying $D$. For Wikivotes, we use the fixed values $\{1,2,4,6\}$ together with the $\{0.60,0.70,0.75,0.80,0.85,0.90,0.95\}$ quantiles of the node-degree distribution; their corresponding integer values appear on the horizontal axis of \Cref{fig:vary-deg}. For Reddit, we use $D\in\{1,2,3,4,6,8\}$, which spans the relevant portion of its degree distribution. This experiment encapsulates the central tradeoff in selecting $D$. On the one hand, a larger degree bound retains more data which reduces projection bias; however, it also increases the sensitivity thereby increasing the amount of noise added.

The plots display this tradeoff through an interior optimum. On Wikivotes, NPGD achieves its best ROC-AUC and accuracy at $D=4$, its lowest test loss at $D=6$, and its best F1 score at $D=26$. Both private methods deteriorate as $D$ becomes large. The non-private projected baseline generally improves with $D$, supporting the conclusion that the decline in the private curves is driven primarily by increasing noise rather than projection bias. On Reddit, the best values are smaller, approximately $D=1$--$2$, and performance declines once $D$ increases beyond this range. This difference is consistent with the datasets' degree distributions; since Reddit has median degree of one, a small $D$ retains much of the useful data and the noise cost soon dominates. Wikivotes has a higher mean degree, making very small values of $D$ more biased, although its larger sample size still permits useful learning after substantial projection. Overall, the results show that good performance can be obtained with a degree bound far below the maximum degree, but that the appropriate value depends on the degree distribution, sample size, and evaluation metric.

\subsection{Computational Efficiency and Scalability}
\label{sec:kashin_compare}
This section details the choices we made when implementing the algorithms to optimize performance and scalability. We also discuss the tradeoffs in runtime, space, and accuracy for different sets of hyperparameters.
\paragraph{Kashin Accuracy-Runtime Tradeoff}
The way we configure the Kashin representation has a significant impact on both predictive performance and runtime. Its principal parameters are the redundancy factor $\lambda=d'/d$ and the uncertainty-principle parameter $\delta$. Increasing $\lambda$ provides more coordinates over which to spread the mass of a vector, reducing the magnitude of its largest coefficient at the cost of operating in a higher-dimensional space. Increasing $\delta$ lowers the clipping threshold in each refinement step, which also produces more uniform coefficients but requires more iterations to reach a fixed reconstruction accuracy. Additional experiments showed that the reconstruction tolerance $\tau$ affects accuracy and runtime as well, but its effect is less central than those of $\lambda$ and $\delta$. We therefore fix $\tau=10^{-6}$ and evaluate every combination of $\lambda\in\{2,3,5\}$ and $\delta\in\{0.1,0.5,0.75,0.95\}$ on Wikivotes.

\begin{figure}[t]
    \centering
    \includegraphics[width=\linewidth]{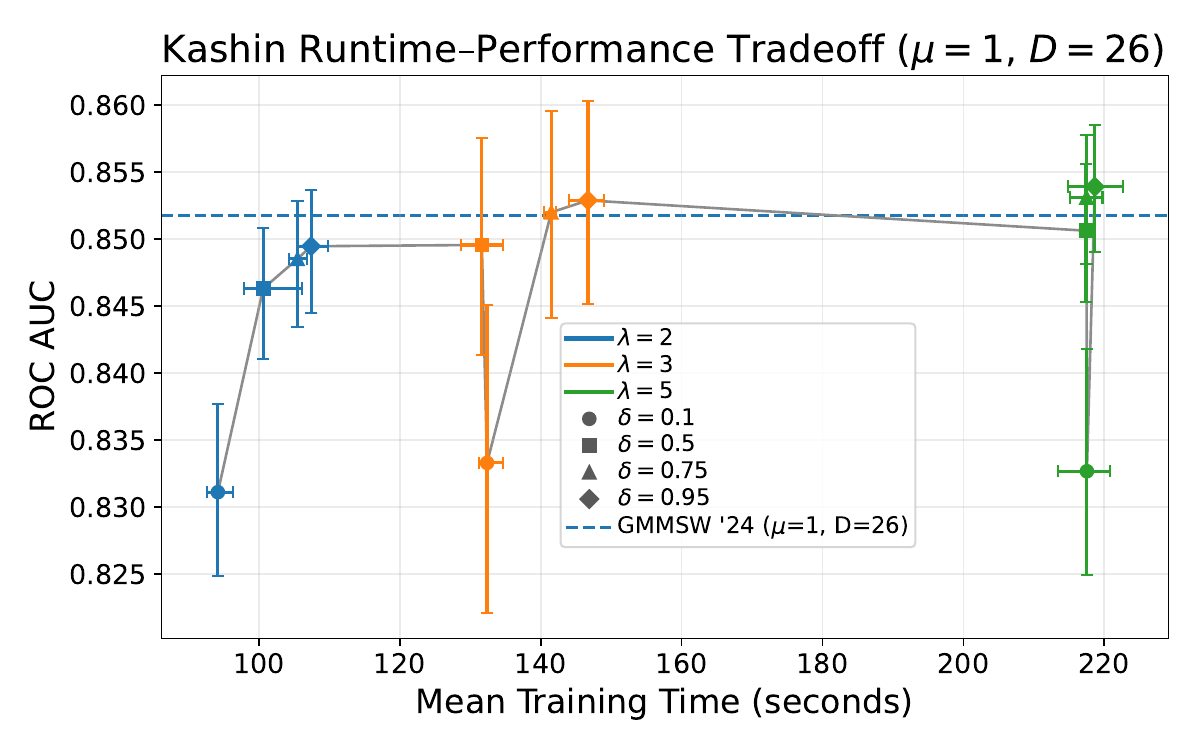}
    \caption{ROC-AUC versus mean training time on Wikivotes for different Kashin representation configurations ($\mu=1$, $D=26$, and $\tau=10^{-6}$). Each combination of $\delta\in\{0.1,0.5,0.75,0.95\}$ (marker) and $\lambda\in\{2,3,5\}$ (color) is evaluated. The dashed line indicates the performance of \gmmsw{} under the same privacy and degree parameters.}
    \label{fig:wikivotes-kashin}
\end{figure}

The results in \Cref{fig:wikivotes-kashin} show that setting $\delta=0.1$ reduces runtime but substantially harms predictive performance for all three values of $\lambda$. Thus, the less aggressive coefficient clipping at $\delta=0.1$ does not spread the gradient mass sufficiently, and its modest computational savings---relative to the $\lambda$ used---do not compensate for the resulting loss in utility. Increasing $\delta$ from $0.5$ to $0.95$ yields smaller but consistent improvements in ROC-AUC for each value of $\lambda$. Increasing $\lambda$ from $2$ to $3$ provides a modest additional improvement, whereas increasing it from $3$ to $5$ produces little further gain while adding roughly $75$ seconds of training time. Overall, moderate-to-large values of $\delta$ improve performance relatively cheaply, while increasing $\lambda$ quickly reaches diminishing returns and is the main driver of runtime. We see this clearly for $\lambda = 3,5$ since the variability in runtime is so small that $\delta=0.5$ is on average slightly faster than $\delta=0.1$, while that is not the case for $\lambda =2$. 

\paragraph{Implementation}

There are several considerations concerning the implementation of our algorithm. The Kashin representation and projection are computationally costly and require access to each per-example gradient, whereas non-private GD can instead differentiate an already aggregated loss. Furthermore, the projection runs at every iteration, and its runtime can vary drastically depending on the implementation. We therefore detail all the important implementation decisions we make in our experiments. We run all Reddit experiments on a shared academic research cluster. A typical job requests one NVIDIA GPU with compute capability of at least 8.0, 16 CPU cores, and 8 GB of memory per core (128 GB in total). Because the cluster is heterogeneous, the scheduler assigns the specific GPU model and CPU architecture according to availability, so these hardware characteristics vary across runs. We initially run the Wikivotes experiments on a personal laptop and later move them to the cluster for efficiency. For the larger model, we perform the computation in batches because of memory limitations; this batching is not represented in the high-level pseudocode. Within each batch, we compute per-example gradients using PyTorch's GPU-accelerated automatic differentiation.

The next step after computing each batch of gradients is converting them to their Kashin representations. This step first requires constructing a tight frame. For Wikivotes, we use the orthonormal matrix obtained from the QR decomposition of a standard Gaussian matrix, which more closely follows the theoretical construction. For the larger Reddit model, this approach is computationally expensive, so we instead use a Sylvester, or Walsh--Hadamard, frame, which supports substantially faster multiplication with less memory overhead. The fast Walsh--Hadamard construction requires a power-of-two width, so the requested transformed dimension $2d=54{,}034$ is increased to $d'=65{,}536=2^{16}$, giving an effective redundancy factor of $\lambda=d'/d\approx2.43$.

Once the frame is constructed, the Kashin algorithm consists primarily of matrix multiplications and coordinate-wise clipping, which can be computed efficiently in batches on a GPU. Although hardware acceleration makes the Kashin representation tractable, it accounts for roughly $60\%$ of the total runtime across the Reddit $\mu$ sweep. Its output at each training iteration is a matrix whose rows are the transformed per-example gradients. For the approximately $42{,}000$ examples in the Reddit training split, this matrix has dimensions $42{,}000\times65{,}536$ and occupies approximately 11 GB. This matrix cannot remain resident in GPU memory alongside the other tensors required for training, so we save it temporarily as a disk-backed matrix; 11 GB therefore represents the peak temporary storage per iteration. A fundamental limitation is the mismatch between how this matrix is produced and how it is later used. While the Kashin representations are computed in batches of rows, the projection takes as input one transformed coordinate, corresponding to one column across the entire training set. All row batches must therefore be completed and stored before the projection can begin, creating an unavoidable synchronization and storage bottleneck in the current implementation.

The projection step then reads the stored Kashin representations and uses them along with the source and target nodes $s$ and $t$ for the projection. Because the Lipschitz extension is applied independently to each transformed coordinate, we can take advantage of parallelization across coordinates. We again utilize a GPU backend, which parallelizes the repeated sorting and edge-selection operations across coordinates. This is another step where, through hardware, we can substantially speed up training time. Nevertheless, the projection must process all $d'$ coordinates at every training iteration. Together with the cost of reading the transformed gradients from disk, the projection accounts for roughly $36\%$ of the total runtime across the Reddit $\mu$ sweep. The relative shares vary across runs because of the heterogeneous hardware, but the Kashin representation and projection together account for approximately $95\%$ of the total runtime.

To summarize runtime, we average the reported training times across the configurations in the $\mu$ sweep. Changing $\mu$ only rescales the Gaussian noise and does not alter the sequence or dimensions of the computations, so it should not produce a meaningful systematic difference in runtime. Averaging across the sweep therefore provides a stable summary that also captures run-to-run variability, including variation in the hardware assigned to the Reddit experiments. On Wikivotes, NPGD takes an average of approximately $238$ seconds ($3$ minutes $58$ seconds), compared with $11.30$ seconds for \gmmsw{}, a $21.06\times$ ratio of mean runtimes. On Reddit, NPGD takes an average of approximately $19{,}068.50$ seconds ($5$ hours $18$ minutes), compared with $527.16$ seconds ($8$ minutes $47$ seconds) for \gmmsw{}, a $36.17\times$ ratio. It is important to note that these ratios compare the observed runtimes of different implementations. Our implementations of \gmmsw{} and the non-private baselines do not use heavily optimized code or specialized hardware.

\section{Discussion}
\label{sec:discussion}

This work demonstrates the feasibility of node-private training as a practical proof of concept. By combining two techniques, we can train models with reasonable error. The algorithm can be configured to prioritize either efficiency or accuracy according to users' preferences. This tradeoff comes from the fact that algorithms that ensure node DP are difficult to design because graph statistics often have high sensitivity. In the worst case, one node can be adjacent to every other node in the graph. Consequently, extensive effort has been devoted to developing and studying techniques for node-private release of various statistics, with approaches typically tailored to the statistic of interest. However, node-DP algorithms for training ML models have received less attention, in part because of concerns about the utility of the resulting model. This work shows that useful node-private models can nevertheless be trained and achieve competitive utility.

Despite providing a practical proof of concept, the algorithm still has several limitations. First, it is restricted to the simple-graph setting rather than the more general hypergraph setting. Whereas a simple graph represents examples attributed to two individuals, a hypergraph allows examples to be attributed to arbitrarily many individuals. The proof of stability for the projection algorithm is valid only for simple graphs, and there is no straightforward extension to hypergraphs. Our algorithm also relies on full-batch gradient descent and does not currently provide the same DP guarantees for stochastic gradient descent. Subsampling introduces randomness that the proof for the deterministic projection cannot yet account for.

This work is intended not only to provide a node-DP algorithm, but also to motivate new approaches to this problem. We believe that removing the need to perform the projection coordinate-wise would be a significant contribution. Working directly with $\ell_2$-bounded vectors is a more natural approach, yet it presents a substantial technical obstacle for the projection we use. Further, performing the projection only once, rather than at every iteration, would greatly improve the efficiency of the algorithm. Although we are aware of algorithmic techniques that avoid performing the projection at every step and coordinate-wise, and on top of this we believe they provide fairly strong privacy, proving their stability remains an open problem. Overall, we believe that more efficient algorithms exist, and we hope this paper serves as a starting point for developing them.

\section*{Open Science}
Along with this paper, we release the code used to run the experiments. This code contains implementations of each algorithm we study, along with configuration files to run the algorithms with different parameters. Instructions are included for downloading datasets and reproducing each plot, though access to accelerated hardware might be necessary for the larger models.

\section*{LLM usage considerations}

LLMs were used to help write code for the paper's experiments and for help with editing the paper. The specific one-dimensional Lipschitz extension we use was suggested by Claude. The overall algorithmic approach and other optimizations were developed by the authors. 

\bibliographystyle{IEEEtran}
\bibliography{references}

\appendices
\crefalias{section}{appendix}

\section{Projection}
\label{appendix:projection}

Below we give the projection algorithm of \asssu{} as Algorithm \ref{alg:dllproject}. We do not claim this as an original contribution of our paper, and we reproduce it here with approval from the authors. We then introduce our algorithm (\ref{alg:splitandsum}) and provide a proof of stability. Ours extends the corresponding proof of \asssu{}, where we explicitly show a tighter lower bound and then apply it to Algorithm \ref{alg:splitandsum}.

\begin{algorithm}[h!]
\caption{$\textsc{Project}(G, D$)}
\label{alg:dllproject}
\begin{algorithmic}[1]
\REQUIRE $G = (V,E,x)$, degree bound $D > 0$
\FOR{$u \in V$}
    \STATE Initialize counter $t(u) \gets 0$
\ENDFOR
\STATE Sort $E$ in decreasing order with respect to edge weights to get ordering $S$
\STATE $\tilde{E} \gets \emptyset$
\STATE $\tilde{G} \gets (V, \tilde{E}, x)$
\FOR{edges $\{u,v\} \in S$}
    \IF{$\max\{t(u), t(v)\} < D$}
        \STATE $\tilde{E} \gets \tilde{E} \cup \{u,v\}$
        \STATE $t(u) \gets t(u) + 1$
        \STATE $t(v) \gets t(v) + 1$
    \ENDIF
\ENDFOR
\RETURN $\tilde{G}$
\end{algorithmic}
\end{algorithm}

\begin{algorithm}[h!]
\caption{$\operatorname{SplitAndSum}(G, D$)}
\label{alg:splitandsum}
\begin{algorithmic}[1]
\REQUIRE $G = (V,E,x)$, degree bound $D > 0$
\STATE $E^{\text{pos}} \gets \{e \in E : x(e) > 0\}$\\
\STATE $E^{\text{neg}} \gets \{e \in E : x(e) < 0\}$\\
\STATE $(V, \widetilde{E}^{\text{pos}}, x) \gets \textsc{Project}((V, {E}^{\text{pos}}, x), D)$
\STATE $(V, \widetilde{E}^{\text{neg}}, x) \gets \textsc{Project}((V, {E}^{\text{neg}}, |x|), D)$
\RETURN $\sum_{e \in \widetilde{E}^{\text{pos}}} x(e) - \sum_{e \in \widetilde{E}^{\text{neg}}} | x(e)|$
\end{algorithmic}
\end{algorithm}

\begin{theorem}
    There is a $D$-Lipschitz extension $f^D_\text{split}$ of $f_\text{sum}$ from $\cgoo_D$ to $\cgoo$ that is computable in time $O(|V| + |E|\log |E|)$
\end{theorem}
\begin{proof}
    For any pair of node-neighboring graphs $G = (V, E, x)$ and $G' = (V', E', x')$ both in $\cgoo$ we claim that 
    \begin{equation}
    \label{eq:stability}
    | f^D_\text{split}(G) - f^D_\text{split}(G') | \leq D,
    \end{equation}
    where $f^D_\text{split}$ is the function computed by Algorithm \ref{alg:splitandsum}.
    
    We first verify that $f^D_\text{split}$ agrees with
$f_\text{sum}$ on the anchor set $\cgoo_D$. Let
$G\in\cgoo_D$. When Algorithm~\ref{alg:dllproject} considers an
edge $\{u,v\}$, at most $D-1$ other edges incident to either
endpoint can have already been added. Thus, neither endpoint is at
capacity, and the edge is retained. This applies to both the positive
and negative subgraphs, so neither projection drops any edges.
Therefore,
\begin{align*}
f^D_\text{split}(G)
&= \sum_{e\in E^{\mathrm{pos}}}x(e)
   - \sum_{e\in E^{\mathrm{neg}}}|x(e)| \\
&= \sum_{e\in E}x(e)
 = f_\text{sum}(G).
\end{align*}
It remains to show that $f^D_\text{split}$ is $D$-Lipschitz on
$\cgoo$.

    To show this, we first start by fixing the two graphs $G$ and $G'$, and assume without loss of generality that $V' \setminus V = \{v^+\}$ and $E' \setminus E = E^+$ where $E^+$ is the set of edges incident to $v^+$. We also assume that $x'(e) = x(e)$ for all $e \in E$ and that $x'(e) = x(e)$ for edges $e \in E^+$, so we can simply use $x$ to denote edge weights in both graphs. 
    %The process by which we will prove \Cref{eq:stability} is to run Algorithm \ref{alg:splitandsum} on both graphs separately, then show their outputs are close. With that in mind, our algorithm also 
    Our algorithm separates the positive and negative edges; doing this for graphs $G$ and $G'$, we get $G_\text{pos}, G_\text{neg}$ and $G_\text{pos}', G_\text{neg}'$ respectively. We focus on the positive graphs for now, but these statements apply exactly to the negative graphs as well, since we take the absolute value of their edge weights.

    \newcommand{\cP}{\mathcal{P}}
    With this setup, note that $G_\text{pos}$ and $G_\text{pos}'$ are still node-neighboring graphs, now with edge weights in $[0,1]$ (same for $G_\text{neg}$ and $G_\text{neg}'$). Therefore, Theorem 4.2 from \cite{AdenAliSSSU26} tells us the absolute value of the difference between $f^D_\text{split}(G_\text{pos})$ and $f^D_\text{split}(G_\text{pos}')$ is bounded above and below by $D$. We will show the stronger statement, 
    \begin{equation}
    \label{eq:strongerbound}
    0 \leq f^D_\text{sum}(G_\text{pos}') - f^D_\text{sum}(G_\text{pos}) \leq D.
    \end{equation}
    
    Define an ordering of edges $\{e_1, \dots, e_k\} \subseteq E^+$ where $x(e_1) \geq x(e_2) \geq \dots \geq x(e_k)$. Next, define the graph $G_i$ to be the same as $G$ but with the edges $e_1, \dots, e_i$ added, for all $i=0, \dots, k$. Notice $G_0 = G$ and $G_k = G'$. We first show the intermediary result that 
    \[
    0 \leq f^D_\text{sum}(G_{\text{pos}, i}) - f^D_\text{sum}(G_{\text{pos}, i -1}')
    \]
    noting that we drop the upper bound since this is already proven in \cite{AdenAliSSSU26}. Using Claim 4.4 from \cite{AdenAliSSSU26}, we know that we have a path $\cP = \{\hat{e}_1, \dots \hat{e}_r\}$ of length $r$ in the symmetric difference graph $\widetilde{E}_{\text{pos}, i} \Delta \widetilde{E}_{\text{pos}, i-1}$. We also have $x(\hat{e}_1) \geq \dots \geq x(\hat{e}_r)$ and that $\{\hat{e}_i \in \cP \colon i \text{ odd}\} \subseteq \widetilde{E}_{\text{pos}, i}$ and $\{\hat{e}_i \in \cP \colon i \text{ even}\} \subseteq \widetilde{E}_{\text{pos}, i-1}$ but $\{\hat{e}_i \in \cP \colon i \text{ odd}\} \cap \{\hat{e}_i \in \cP \colon i \text{ even}\} = \emptyset$. 

    Our goal is to lower bound the difference 
    \[
    \sum_{e \in \widetilde{E}_{\text{pos}, i}} x(e) - \sum_{e \in \widetilde{E}_{\text{pos}, i-1}} x(e).
    \]
    Given Claim 4.4 in \cite{AdenAliSSSU26}, this reduces to taking the difference between the sum of odd indexed edges in the path and even indexed edges in the path. This is simple due to their non-increasing ordering. We can pair up each edge, and find that their difference will be small, but greater than 0, so the whole difference must be greater than 0. Starting with the case where $r$ is even,  
    \begin{align*}
    &\sum_{e \in \widetilde{E}_{\text{pos}, i}} x(e) - \sum_{e \in \widetilde{E}_{\text{pos}, i-1}} x(e) \\
    &\qquad = (x(\hat{e}_1) - x(\hat{e}_2)) + \dots + (x(\hat{e}_{r-1}) - x(\hat{e}_{r})) \\
    &\qquad \geq 0 + \dots + 0 \geq 0. 
\end{align*}
In the case where $r$ is odd, we have one term left over, 
\begin{align*}
    &\sum_{e \in \widetilde{E}_{\text{pos}, i}} x(e) - \sum_{e \in \widetilde{E}_{\text{pos}, i-1}} x(e) \\
    &\qquad = (x(\hat{e}_1) - x(\hat{e}_2)) + \dots + (x(\hat{e}_{r-2}) - x(\hat{e}_{r-1})) + x(\hat{e}_r) \\
    &\qquad \geq 0 + \dots + 0 + x(\hat{e}_r) \geq 0,
\end{align*}
    and the bound still holds. 

    Knowing that $f^D_\text{sum}(G_{\text{pos}, i}) \geq f^D_\text{sum}(G_{\text{pos}, i-1})$ for all $i=1, \dots k$, we can obtain the bound in \Cref{eq:strongerbound} by observing 
    \begin{align*}
    f^D_\text{sum}(G_{\text{pos}}') = f^D_\text{sum}(G_{\text{pos}, k}) &\geq f^D_\text{sum}(G_{\text{pos}, k-1})\\
    &\geq \dots \geq f^D_\text{sum}(G_{\text{pos}, 0}) =  f^D_\text{sum}(G_{\text{pos}}).
    \end{align*}

    Having shown the monotonicity of the extension, we use this property to prove the final bound of \Cref{eq:stability}. To prove it, we first start by rewriting the difference as 
    \begin{align*}
     &\left| f^D_\text{split}(G') - f^D_\text{split}(G_\text{pos}) \right| = \left| \sum_{e \in \widetilde{E}'} x(e) - \sum_{e \in \widetilde{E}} x(e)\right| \\
     &= \left| \left(\sum_{e \in \tilde{E}'_\text{pos}} x(e) - \sum_{e \in \tilde{E}'_\text{neg}} |x(e)| \right)  -  \left(\sum_{e \in \tilde{E}_\text{pos}} x(e) - \sum_{e \in \tilde{E}_\text{neg}} |x(e)| \right) \right|,
    \end{align*}
    which comes from running Algorithm \ref{alg:splitandsum} on both $G$ and $G'$. Now using \Cref{thrm:extension} in combination with the monotonicity of the extension, we know that 
    \[
    0 \leq \sum_{e \in \tilde{E}'_\text{pos}} x(e) - \sum_{e \in \tilde{E}_\text{pos}} x(e) \leq D
    \]
    and 
    \[
    0 \leq \sum_{e \in \tilde{E}'_\text{neg}} |x(e)| - \sum_{e \in \tilde{E}_\text{neg}} |x(e)| \leq D.
    \]
    So, the sum of $x(e)$ over the kept positive edges goes up by some value $0 \leq a \leq D$, and the sum of the absolute value $|x(e)|$ over the kept negative edges goes up by some $0 \leq b \leq D$. Therefore,
    \[
    | f^D_\text{split}(G') - f^D_\text{split}(G) | = | a - b |  \leq D.
    \]

    For the running time, splitting the edges takes $O(|E|)$ time, and each of the two calls to Algorithm~\ref{alg:dllproject} takes $O(|V| + |E|\log |E|)$ time.

\end{proof}

\section{Kashin Representation}
\label{appendix:kashin-rep}
Below we provide pseudocode for the algorithm we implement to construct our Kashin representation. The first is an algorithm for constructing a tight frame, the one we use in our Wikivotes experiments. The second is the actual Kashin representation algorithm of \cite{UPandvectorquantization}.

\begin{algorithm}[h!]
  \caption{Tight Frame Constructor}
  \label{alg:tightframeconstructor}
  \begin{algorithmic}[1]
    \REQUIRE Original dimension $d$, Kashin dimension scaling factor $\lambda$
    \ENSURE A tight frame matrix $U \in \mathbb{R}^{d \times d'}$ satisfying $UU^T = I_d$
    \STATE $d' \gets \lambda d$
    \STATE Sample $G \in \mathbb{R}^{d' \times d}$ with entries $G_{ij} \sim \mathcal{N}(0,1)$
    \STATE Compute the reduced QR decomposition $G = QR$, where
           $Q \in \mathbb{R}^{d' \times d}$ has orthonormal columns and
           $R \in \mathbb{R}^{d \times d}$ is upper triangular
    \STATE $U \gets Q^T$
    \RETURN $U$
  \end{algorithmic}
\end{algorithm}

\begin{algorithm}[h!]
  \caption{Computing Kashin's representation}
  \label{alg:kashinrep}
  \begin{algorithmic}[1]
    \REQUIRE Parseval frame $U \in \mathbb{R}^{d \times d'}$ satisfying
             $UU^\top = I_d$, $\delta \in (0,1)$,
             vector $x \in \mathbb{R}^d$, maximum iterations $r$,
             and tolerance $\tau > 0$
    \ENSURE Kashin coefficients $a \in \mathbb{R}^{d'}$
    \STATE $a \gets 0 \in \mathbb{R}^{d'}$
    \STATE $z \gets x$
    \FOR{$t = 1$ \TO $r$}
      \IF{$\lVert z \rVert_2 < \tau$}
        \STATE \textbf{break}
      \ENDIF
      \STATE $b \gets U^\top z$
      \STATE $M \gets \lVert z \rVert_2 / \sqrt{\delta d'}$
      \STATE $\widehat{b} \gets
        \operatorname{sign}(b)\odot\min\{|b|,M\}$
        \COMMENT{element-wise clipping}
      \STATE $a \gets a+\widehat{b}$
      \STATE $z \gets z-U\widehat{b}$
    \ENDFOR
    \RETURN $a$
  \end{algorithmic}
\end{algorithm}

\subsection{Additional Kashin Experiments}
\label{appendix:kashin-experiments}

\begin{figure}[h!]
    \centering
    \includegraphics[width=\linewidth]{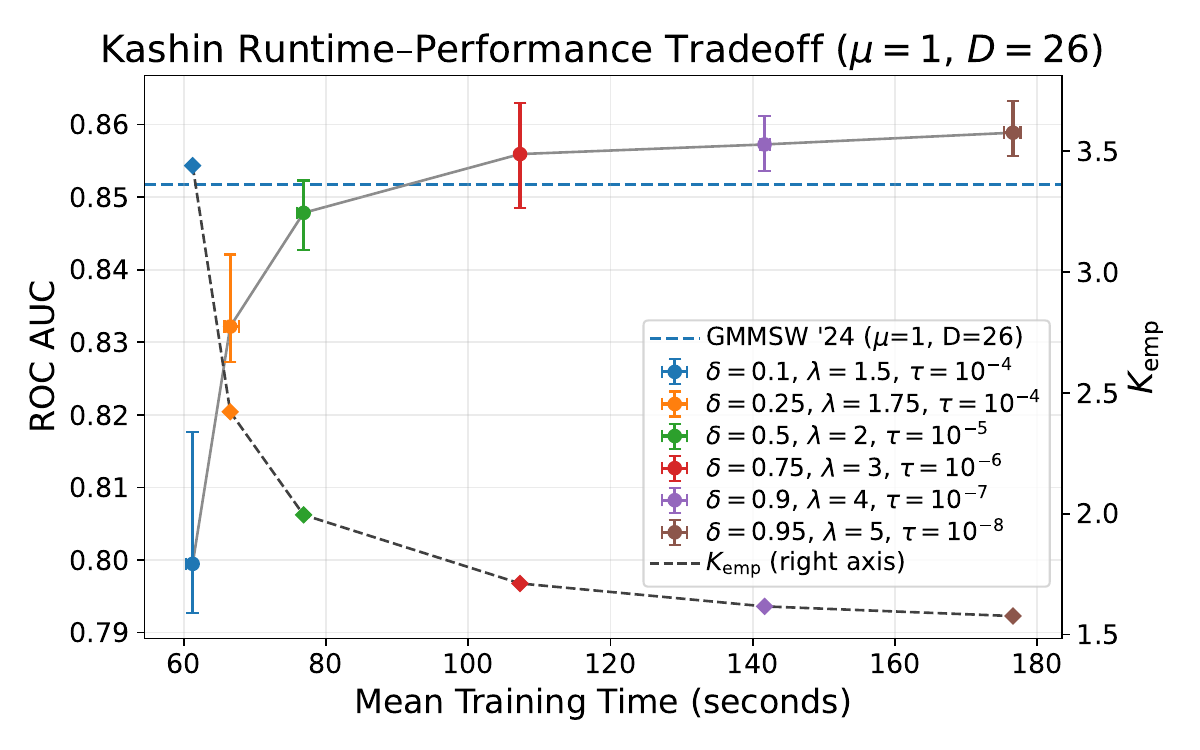}
    \caption{ROC-AUC (left axis) and $K_\text{emp}$ (right axis) versus mean training time on Wikivotes ($\mu=1$, $D=26$) for Kashin configurations in which $\delta$, $\lambda$, and $\tau$ are varied jointly. The dashed horizontal line indicates the performance of \gmmsw{} under the same privacy and degree parameters.}
    \label{fig:wikivotes-kashin-kemp}
\end{figure}

\Cref{fig:wikivotes-kashin-kemp} shows further Kashin experiments on Wikivotes in which we varied $\delta$, $\lambda$, and $\tau$ simultaneously, reporting ROC-AUC, mean training time, and the resulting $K_\text{emp}$ for each configuration.

\end{document}